\documentclass[conference]{IEEEtran} 

\usepackage{amsmath, amssymb}
\usepackage{textcomp}
\usepackage{mathrsfs}
\usepackage{booktabs}
\usepackage{bm}
\usepackage{xcolor,varwidth}
\usepackage{multicol}
\usepackage{graphics,graphicx}
\usepackage{subfig}

\def\mbW{\mathbf{W}}
\def\mbTheta{{\bm\Theta}}
\def\mbtheta{{\bm\theta}}
\def\mbX{\mathbf{X}}
\def\mbM{\mathbf{M}}
\def\mbY{\mathbf{Y}}
\def\mbS{\mathbf{S}}
\def\RR{\mathbb{R}}

\begin{document}

\title{Machine Learning in Appearance-based Robot Self-localization}

\author{\IEEEauthorblockN{Alexander Kuleshov\IEEEauthorrefmark{1}, 
Alexander Bernstein\IEEEauthorrefmark{1}\textsuperscript{,}\IEEEauthorrefmark{2},
Evgeny Burnaev\IEEEauthorrefmark{1}\textsuperscript{,}\IEEEauthorrefmark{2},
Yury Yanovich\IEEEauthorrefmark{2}}

\IEEEauthorblockA{\IEEEauthorrefmark{1}Skolkovo Institute of Science and Technology, Moscow, Russia\\
Emails: A.Kuleshov@skoltech.ru, A.Bernstein@skoltech.ru, E.Burnaev@skoltech.ru, \\ \IEEEauthorrefmark{2} Kharkevich Institute for Information Transmission Problems RAS, Moscow, Russia \\
Email: Yury.Yanovich@iitp.ru}
}

%
%

\maketitle              

\begin{abstract} An appearance-based robot self-localization problem is considered in the machine learning framework. The appearance space is composed of all possible images, which can be captured by a robot's visual system under all robot localizations. Using recent manifold learning and deep learning techniques, we propose a new geometrically motivated solution based on training data consisting of a finite set of images captured in known locations of the robot. The solution includes estimation of the robot localization mapping from the appearance space to the robot localization space, as well as estimation of the inverse mapping for modeling visual image features. The latter allows solving the robot localization problem as the Kalman filtering problem.
\end{abstract}

\begin{IEEEkeywords}
	machine learning; mobile robot self-localization; appearance-based learning; deep learning; manifold learning
\end{IEEEkeywords}
\section{Introduction}
\label{sec1}

Machine learning is an essential and ubiquitous framework for solving a wide range of tasks in different application areas. One of these tasks is the problem of estimating position (localization) of a mobile robot moving in an uncertain environment, which is necessary for understanding the environment to make navigational decisions. Usually to solve self-localization problem for navigation of autonomous robots a built in visual sensor system (camera) is utilized.

The most common and basic method for performing localization is through dead-reckoning using data received from odometer sensors. This technique integrates a history of sensor readings and executed actions (e.g., velocity history) of the robot over time to determine changes in positions from its starting location \cite{bib1,bib2,bib3}. It is common to combine the additional localization technique based on visual data such as images or range profiles (see review \cite{bib4}) with dead reckoning applying the extended Kalman filter to probabilistically update the robot position \cite{bib5,bib6}. In passive vision-based localization, position of an autonomous mobile robot equipped by a visual system (e.g., omnidirectional imaging system \cite{bib7,bib8,bib9} or camera with steerable orientation \cite{bib10}) can be estimated from images captured by the visual system. Continuous set of possible images, which can be captured by the visual system under all possible image formation parameters (relative position and orientation of the robot moving in a certain workspace, as well as camera intrinsic parameters and illumination function) is called Appearance Space (AS). 

In this paper, we consider the most popular case when the AS is parameterized by robot localization  consisting only of its position and orientation. Assuming that captured images allow distinguishing and recognizing localizations from which they have been taken, the solution to the considered appearance-based (passive vision-based) robot localization problem is a mapping from the AS to the Localization Space (LS), formed by all possible robot localizations.

We consider the appearance-based localization problem in the machine learning framework: the appearance-based model, which describes the relation between observed images and robot locations, is constructed using a finite set of captured images taken in known positions. This model allows estimating an unknown robot localization from a newly acquired image. Such appearance-based learning framework has become very popular in the field of robot learning \cite{bib11}.

Various appearance-based models (aka maps), describing the underlying low-dimensional structure in the AS, are usually constructed from training positions-images data using supervised learning techniques. Such models provide internal representation of the high-dimensional AS by certain visual features (or landmarks) extracted from images, see a short review in \cite{bib12}.

In many appearance-based methods input images from the AS are considered holistically, in relation to other images, and natural visual features are computed by projecting images onto low-dimensional subspaces [9, 13, 14] usually via the Principal Component Analysis (PCA). Various types of regression between images or their low-dimensional features and robot coordinates are constructed including Gaussian process regression, random forest, etc. \cite{bib15,bib16,bib17,bib18,bib19,burn1,burn2,burn3}.

This paper proposes new geometrically motivated machine learning approach to the appearance-based robot localization problem by combining a few advanced techniques.

At first a deep learning model extracts visual features, representing a learned mapping from the AS to an image Visual Feature Space (VFS) (image of the AS under this mapping). The VFS is a low-dimensional manifold (surface) in the ambient high dimensional space with intrinsic dimensionality equal to the dimensionality of the LS, which in turn is equal to three. Thus, the solution of the Robot localization problem reduces to the regression problem with manifold valued inputs, ``living'' in the VFS, and outputs, belonging to the LS. 

Next the manifold learning procedure from \cite{bib20} is used to construct regression with high-dimensional manifold valued inputs.

After that inverse mapping from the LS to the VFS is constructed via special nonlinear dimensionality reduction procedure \cite{bib21,bib22,bib23}. This mapping allows predicting visual features of an image that will be captured at the predicted localization of the mobile robot based on chosen navigational decisions. The latter allows solving the robot localization problem as the Kalman filtering problem.

The paper is organized as follows. Section \ref{sec2} contains rigorous statement of the appearance-based robot localization problem. Proposed approach is described in Section \ref{sec3}. Section \ref{sec4} provides details of this solution. Section \ref{sec5} describes results of numerical experiments. Conclusions are given in section \ref{sec6}.

\section{Robot Localization: Rigorous Problem Statement}
\label{sec2}

Let a mobile robot, equipped with a visual system (for example, an omnidirectional imaging system), moves on a 2D-workspace. Its localization $\mbtheta = (\mbtheta_{RC},\mbtheta_{RO})\in\RR^3$ is a three-dimensional vector consisting of \textbf{R}obot \textbf{P}osition $\mbtheta_{RP} \in \RR^2$ (robot coordinates in a $2D$-workspace) and \textbf{R}obot \textbf{O}rientation (an angle) $\mbtheta_{RO}\in\RR^1$ relative to the coordinate system in the workspace. Let us denote by $\mbTheta\subset\RR^3$ a subset consisting of all possible robot localization parameters and called Localization space. 

Let an image, captured by the robot imaging system, consists of $p$ pixels and, thereby, is represented by an image-vector $X \in \RR^p$. We denote by $X =\varphi(\mbtheta)\in\RR^p$ an image, captured by the robot with the localization parameter $\mbtheta$, which is described by the Image modeling function $\varphi$ with the domain of definition $\mbTheta$. Let us denote by
\begin{equation}
\label{eq1}
\mbX = \varphi(\mbTheta) = \{X = \varphi(\mbtheta),\,\mbtheta\in\mbTheta\subset\RR^3\}\subset\RR^p
\end{equation}
the Appearance space consisting of all possible images, which can be captured by the mobile robot and parameterized by the robot localization parameter $\mbtheta\in\RR^3$. 

Assuming that images, captured by the robot in different localizations, are different, the Image modeling function $\varphi:\,\mbTheta\to\mbX$ is a one-to-one mapping from the LS to the AS. Thus, the AS $\mbX$ is a three-dimensional manifold without self-intersections (Appearance manifold, AM), parameterized by the chart $\varphi$ and  embedded in the ambient $p$-dimensional space. Therefore, there exists an inverse mapping
\begin{equation}
\label{eq2}
	\psi = \varphi^{-1}:\,X\in\mbX\to\mbTheta = \psi(X) \in \mbTheta,
\end{equation}
called Localization function from the AM $\mbX$ to the LS $\mbTheta$.

The functions $\varphi$ and $\psi$, as well the AM $\mbX$, are unknown, and the Robot localization problem consists in constructing the robot localization $\mbtheta = \psi(X)$ from the image $X = \varphi(\mbtheta) \in \mbX$. We consider this problem in the machine learning framework. Let us denote by 
\begin{equation}
\label{eq3}
	\mbS_{\mathbf{X},\mbtheta} = \{(X_i,\mbtheta_i), i = 1, 2, \ldots, n\} 	
\end{equation}
a training dataset consisting of images $\{X_i = \varphi(\mbtheta_i)\}$, captured by the robot visual system in known localizations $\{\mbtheta_i \in \mbTheta\}$ when robot moves in the workspace randomly or on a regular grid. For example, the mobile robot, described in \cite{bib7}, captures omnidirectional images every $25$ centimeters along robot random paths; reference positions are located on a regular grid with cells of size either $25$cm in a $2.7$m$\times5.4$m workspace \cite{bib8} or $1$m in a $20$m$\times20$m workspace \cite{bib9}, respectively. 

We consider the robot localization problem as Localization function estimation problem: to recover an unknown Localization function $\mbtheta = \psi(X)$ at an arbitrary out-of-sample point $X \in \mbX$, from its known values $\{\mbtheta_i = \psi(X_i)\}$ at known points $\{X_i\}$.

This problem is a regression one with high-dimensional manifold valued inputs. For example, input dimensionality $p$ equals $16384$, $10240$, and $3925$ in case of panoramic images considered in \cite{bib7,bib8,bib9}, respectively; $p = 6912$ and $p = 4096$ for two examples considered in \cite{bib10}.

\section{Robot Localization: Proposed Approach}
\label{sec3}

Proposed approach consists of four successively executed steps:
\begin{itemize}
	\item
    	pre-process captured images to get image visual features;
     \item
     	estimate robot localization from visual features of captured images;
     \item
     	predict visual features of an image that will be captured at a new given robot localization;
     \item
     	solve the robot localization problem via the Kalman filtering technique.
\end{itemize}
Details are given below.

\subsection{Preprocessing Captured Images}
\label{sec3.1}
Image data requires subject-matter expertise to extract key features. Deep convolutional neural networks (DCNNs) extract features automatically from domain-specific images, without any feature engineering techniques. This process makes DCNNs suitable for extraction of visual features:
\begin{itemize}
\item DCNN with many layers is trained using some extensive image database,
\item Usually initial layers learn low-order features (e.g. color, edges, etc.),
\item Final layers learn higher order representations (specific to input image) that are subsequently used for image classification and transfer learning, see \cite{trans2}.
\end{itemize}

The constructed mapping, realized by final layers of DCNN, determines Visual Feature Space (VFS)
\begin{equation}
  \label{eq4}
  \mbW = \omega(\mbX) = \{w =\omega(X), X \in \mbX\} \subset \RR^m,
\end{equation}
consisting of visual features for all images from the AS.

As well the AS $\mbX$, the VFS $\mbW = \omega(\varphi(\mbTheta))$ is a three-dimensional manifold parameterized by the chart $w = \Phi(\mbtheta)  \equiv \omega(\varphi(\mbtheta))$ and embedded in an ambient $m$-dimensional space. The mapping $\Phi(\mbtheta)$, called Feature modeling function, predicts features $w = \omega(X)$ of an image $X = \varphi(\mbtheta)$ that is captured at an arbitrary robot localization $\mbtheta \in \mbTheta$.

\subsection{Nonlinear Regression on High-dimensional Inputs: Localization Function Estimation}
\label{sec3.2}

Using the constructed mapping $\omega$, initial Localization function estimation problem is reduced to the regression problem on the VFS $\mbW$: using dataset 
\begin{equation}
  \label{eq5}
  S_{\mathbf{W},\mbtheta} = \{(w_i, \mbtheta_i), i=1,2, \dots, n\}
\end{equation}
with known robot localizations $\{\mbtheta_i = V(w_i)\}$ and computed features $\{w_i = \omega(X_i)\}$ from  images $\{X_i\}$, captured in these localizations, we estimate an unknown robot position $\mbtheta = V(w)$ from visual features $\omega(X)$ of an image $X = \varphi(\mbtheta)$, captured at some new robot localization $\mbtheta$. The solution $V(w)$ determines the sought-for mapping $\psi(X) = V(\omega(X)))$.

Initial and reduced regression problems deal with high dimensional inputs, and standard regression methods perform poorly due the statistical and computational `curse of dimensionality' phenomenon. However, by construction the input spaces in both problems are three-dimensional manifolds parameterized by the charts $\varphi$  and $\omega(\varphi)$ and embedded in the ambient high-dimensional spaces $\mbX$ and $\mbW$, respectively. Taking into account this fact, in order to avoid the curse of dimensionality phenomena most of approaches use various dimensionality reduction techniques (usually, PCA) for constructing low-dimensional visual features by projecting captured images onto low-dimensional linear subspaces. The robot localization function is estimated from such visual features \cite{bib11,bib13,bib14}. However, constructed low-dimensional linear subspaces may have a larger dimension than the real intrinsic dimension, which is equal to three.

Our approach uses new solution to nonlinear multi-output regression problem with input belonging to unknown manifold  \cite{bib20}. The core of the solution is based on a special nonlinear dimensionality reduction method called Grassmann \& Stiefel Eigenmaps \cite{bib21,bib22}; this method has already been successfully used to solve certain statistical learning problem \cite{bib23}.

The solution \cite{bib20} results in 
\begin{itemize}
	\item embedding mapping $y = h(w)$ from the VFS $\mbW$ to the three-dimensional Representation Space (RS) $\mbY = h(\mbW)$ consisting of representations $y = h(w) \in \RR^3$ for computed features $w \in \mbW$;
    \item recovered mapping $\mbtheta= g_{\mbTheta}(y)$, which maps the low-dimensional representation $y = h(w)$ of visual features $w = \Phi(\mbtheta)$ to the robot localization $\mbtheta$. 
\end{itemize}

Therefore, the mapping $V(w)$ is estimated by the statistic $V^*(w) = g_{\mbTheta}(h(w))$ and sought-for Localization function is estimated by

\begin{equation}
\label{eq6}
	\mbtheta = \psi^*(X) = V^*(\omega(X)) = g_{\mbTheta}(h(\omega(X))).	
\end{equation}

\subsection{Nonlinear Regression with High-dimensional Inputs: Feature modeling function estimation}

Consider the problem of Feature modeling function $\Phi(\mbtheta)$ estimation from the dataset $S_{\mathbf{W},\mbtheta}$ \eqref{eq5}. This regression problem can be described as follows: estimate an unknown mapping $\Phi(\mbtheta)$ from the LS $\mbTheta$ to the VFS $\mbW$ using known visual features $\{\Phi(X_i) = w_i = \omega(X_i)\}$ of images $\{X_i = \varphi(\mbtheta_i)\}$ captured at known points $\{\mbtheta_i\}$. Here we have a regression problem with high-dimensional manifold valued outputs belonging to the $3$-dimensional manifold $\mbW$ \eqref{eq4}.

The solution \cite{bib20} also realizes recovery mapping $w = g_W(y)$ from the low-dimensional representation $y = h(w) \in \mbY$ to its preimage (visual features $w \in \mbW$). This mapping is `approximately inverse' to the embedding mapping $h$ and provides proximity

\begin{equation}
\label{eq7}
	g_W(h(w)) \approx w \quad\,\,\,\,\text{for all} \;\; w \in \mbW.
\end{equation}

This quantity allows estimating the Feature modeling function by the formula

\begin{equation}
\label{eq8}
	w = \Phi^*(\mbtheta) = g_W(h(\Phi(\mbtheta))).	
\end{equation}

\subsection{Kalman Filtering in Robot Localization Problem}

The constructed estimator $\Phi^*(\mbtheta)$ \eqref{eq8}, which predicts  visual features $w = \omega(X)$ of an image $X = \varphi(\theta)$, captured at the localization $\mbtheta \in \mbTheta$, can be used in a Kalman filtering procedure \cite{bib5} for robot localization.

Robot navigation consists in choosing the control $U(t)$ at given time moments $t = 0, 1, 2, \dots$. Let $\mbtheta(t)$ be a current robot position at time $t$, then, under some chosen control $U(t)$, the robot must move to the expected position
\begin{equation}
\label{eq9}
	\mbtheta(t+1) = F(\mbtheta(t), U(t)) \equiv F_t(\mbtheta(t)),	
\end{equation}
where $F(\mbtheta, U)$ is a known function defined by a solution of a navigation motion control problem.

In practice, the estimated position $\mbtheta_t(t)$ of the robot at time $t$ is only known, which differs from the exact position $\mbtheta(t)$; the exact position $\mbtheta(t+1)$ at time $(t+1)$ also differs from the expected position $\mbtheta_t(t+1) = F_t(\mbtheta_t(t))$.

Let a robot visual sensing system provides a captured image $X(t+1) = \varphi(\mbtheta(t+1))$ at the moment $(t+1)$. We want to solve a filtering problem to improve the predicted localization $\mbtheta_t(t+1)$ from the captured image $X(t+1)$.

The constructed estimator $\Phi^*(\mbtheta)$ \eqref{eq8} allows predicting $w^*(t+1) = \Phi^*(\mbtheta_t(t+1))$ for visual features $w(t+1) = \omega(X(t+1))$ of the captured image $X(t+1)$, and the standard Kalman filter \cite{bib5} constructs the improved localization $\mbtheta_{t+1}(t+1)$ as
\begin{equation*}
	\mbtheta_{t+1}(t+1) = \mbtheta_t(t+1) + B(t+1)\times(w(t+1) - w^*(t+1)),
\end{equation*}    
where $B(t+1)$ is a Kalman gain.

Using the estimator $\psi^*(X)$ \eqref{eq6} for the Localization function $\psi(X)$ \eqref{eq2}, we can use a quantity $\psi^*(X(t+1))$, representing visual features, as an estimator of the robot pose in which the image $X(t+1)$ has been taken, and construct the estimator $\mbtheta_{t+1}(t+1)$ as
\begin{eqnarray*}
	&\mbtheta_{t+1}(t+1) = \mbtheta_t(t+1) + b(t+1)\times \\ & \;\;\;\;\;\;\;\;
    \times(\psi^*(X(t+1)) - \psi^*(X^*(t+1))),
\end{eqnarray*} 
where $b(t+1)$ is another gain function.

As measurements, used in filtering procedures, it is possible to use low-dimensional representations $y = h(w)$ of visual features $w = \omega(X)$, obtained as a solution of nonlinear multi-output regression on unknown input manifold problem \cite{bib20}. Such estimator has the form 
\begin{eqnarray*}
	& \mbtheta_{t+1}(t+1) = \mbtheta_t(t+1) + b(t+1)\times\\ 
     & \;\;\;\;\;\;\;\; \times(h(\omega(X(t+1))) - h(w^*(t+1))),
\end{eqnarray*}    
where $b(t+1)$ is some gain function.  

For choosing the optimal gain functions in above Kalman filtering procedures, it is necessary to know covariance matrices for deviations between observations and their expected values, as well as between the expected robot localizations $\mbtheta_t(t+1) = F_t(\mbtheta_t(t))$ \eqref{eq9} and its real pose $\mbtheta(t+1)$. Corresponding covariance matrices can be estimated from samples $\mbS_{\mathbf{X},\mbtheta}$ \eqref{eq3} and $S_{\mathbf{W},\mbtheta}$ \eqref{eq5} in which robot localizations are known with high accuracy.

\section{Robot Localization: Solution}
\label{sec4}

Preprocessing step is a standard one and usage of Kalman filtering procedures in robot localization problem were described in Sections \ref{sec3.1} and \ref{sec3.2}, respectively. In this section we describe shortly the estimating procedures for Robot localization and Feature modeling functions based on the solution of nonlinear regression problem with high-dimensional manifold valued inputs.

\subsection{Robot Localization Manifold Estimation Problem}

Consider an unknown smooth manifold called Regression manifold (RM)
\begin{equation}
\label{eq10}
	\mbM = \{Z=F(\mbtheta),\,\mbtheta\in\mbTheta\subset\RR^3\}\subset\RR^{m+3}
\end{equation}
with the intrinsic dimension $q = 3$, which is embedded in an ambient $(m+3)$-dimensional Euclidean space and parameterized by an unknown chart
\begin{equation}
\label{eq11}
F:\,\mbtheta\in\mbTheta\subset\RR^3\to Z = F(\mbtheta) = 
\left(
\begin{array}{c}
\Phi(\mbtheta)\\
\mbtheta
\end{array}
\right)
\in\mbM,
\end{equation}    
defined on the Localization space $\mbTheta$. 
The RM $\mbM$ can be considered as a direct product of the VFS $\mbW = \Phi(\mbTheta)$ and the LS $\mbTheta$.

Let 
\begin{equation}
\label{eq12}
J_F(\mbtheta) = \nabla_{\mbtheta}F(\mbtheta) = 
\left(
\begin{array}{c}
J_{\Phi}(\mbtheta)\\
\mathbf{I}_3
\end{array}
\right),
\end{equation}
be $(m+3)\times 3$ Jacobian matrix of the mapping $F$ \eqref{eq11} which is split into $m\times 3$ Jacobian matrix $J_{\Phi}(\mbtheta)$ of the mapping $\Phi(\mbtheta) = \omega(\varphi(\mbtheta))$ and $\mathbf{I}_3$ being a $3\times3$ unit matrix. The Jacobian $J_F(\mbtheta)$ \eqref{eq12} determines a three-dimensional linear space $L(Z) = \mathrm{Span}(J_F(\mbtheta))$ in $\RR^{m+3}$ which is a tangent space to the RM $\mbM$ at the point $Z = F(\mbtheta) \in\mbM$; hereinafter, $\mathrm{Span}(H)$ is a linear space spanned by columns of an arbitrary matrix $H$.

The set $\mathrm{TB}(\mbM) = \{(Z, L(Z)):\, Z \in \mbM\}$ consists of points $Z$ of the RM $\mbM$, equipped by tangent spaces $L_F(X)$ at these points, is known in the Manifold theory \cite{bib25} as the Tangent Bundle of the RM $\mbM$.

The dataset $\mbS_{\mathbf{W},\mbtheta}$ \eqref{eq5}, written in the form
\begin{align}
    \mbS_{\mathbf{W},\mbtheta} 
    &= \left\{Z_i=
    \left(
    \begin{array}{c}
    X_i = \Phi(\mbtheta_i)\\
    \mbtheta_i
    \end{array}
    \right), i = 1,2,\ldots,n\right\}=\notag\\
    &\left\{Z_i=
    \left(
    \begin{array}{c}
    w_i\\
    \mbtheta_i = V(w_i)
    \end{array}
    \right), i = 1,2,\ldots,n\right\}
    	\label{eq13}
\end{align}
can be considered as a sample from the unknown RM $\mbM$ \eqref{eq4}.

Let us consider certain dimensionality reduction problem called Tangent bundle manifold learning problem \cite{bib26} for the RM $\mbM$: estimate the Tangent Bundle $\mathrm{TB}(\mbM)$ given the sample $\mbS_{\mathbf{W},\mbtheta}$ \eqref{eq13} from the unknown RM $\mbM$.

\subsection{Robot Localization Manifold Estimation: GSE Solution}
\label{sec4.2}

Using Grassmann \& Stiefel Eigenmaps (GSE) method \cite{bib21,bib22}, applied to the sample $S_{\mathbf{W},\mbtheta}$ \eqref{eq13}, we construct the solution to the Tangent bundle manifold learning problem, resulting in the following quantities: 
\begin{itemize}
	\item 
    	sample-based area $\mbM^* \subset \RR^{m+3}$ which is close to the unknown RM $\mbM$, 
      \item 
      	embedding mapping $h_{GSE}(Z)$ from the area $\mbM^*$ to the Representation Space (RS) $\mbY_{GSE} = h_{GSE}(\mbM^*)\subset\RR^3,$ 
  \item 
  	recovery mapping $g_{GSE}(y)$ from the RS $\mbY_{GSE}$ to $\RR^{m+3}$, 
  \item 
  	$(m+3)\times 3$ matrix $G_{GSE,g}(y)$ defined on the RS $\mbY_{GSE}$, 
\end{itemize}
which together provides both 
\begin{itemize}
\item proximity
\begin{equation}
\label{eq14}
	Z_{GSE}(Z) \equiv g_{GSE}(h_{GSE}(Z)) \approx Z\,\,\mbox{for all }\,\,Z \in \mbM^*,
\end{equation}
between initial and recovered points $Z$ and $Z_{GSE}(Z)$. Thanks to \eqref{eq14} we get small Hausdorff distance $d_H(\mbM, \mbM_{GSE})$ between the RM $\mbM$ and the three-dimensional recovered regression manifold (RRM)
\begin{equation}
\label{eq15}
	\mbM_{GSE} = \{g_{GSE,g}(y) \in \RR^{m+3}: y \in \mbY_{GSE}\subset\RR^3\},
 \end{equation}   
embedded in the ambient $(m+3)$-dimensional Euclidean space; 
\item proximity
\begin{equation}
\label{eq16}
	G_{GSE,g}(y) \approx J_{GSE,g}(y)\,\,\mbox{ for all }\,\, y \in \mbY_{GSE},
\end{equation}
in which $J_{GSE,g}(y)$ is a Jacobian matrix of the mapping $g_{GSE}(y)$. Thanks to \eqref{eq16} we get proximity between the tangent space $L(Z)$ to the RM $\mbM$ at the point $Z$ and the tangent space $L_{GSE}(Z) = \mathrm{Span}(G_{GSE,g}(h_{GSE}(Z)))$ to the RRM $\mbM_{GSE}$ \eqref{eq15} at the nearby recovered point $Z_{GSE}(Z)$. The proximity between these tangent spaces, considered as elements of the Grassmann manifold, is defined using chosen metric on the Grassmann manifold.
\end{itemize}

Therefore, the tangent bundle 
\[
	\mathrm{TB}(\mbM_{GSE}) = \{(Z_{GSE}(Z), L_{GSE}(Z)):\, Z \in\mbM_{GSE}\}
\]
of the RRM $\mbM_{GSE}$ accurately approximates the tangent bundle $\mathrm{TB}(\mbM)$.

Note also that the original GSE algorithm \cite{bib21,bib22} has computational complexity $O(n^3)$ for a sample of size $n$; the incremental version of the GSE \cite{bib27} has significantly smaller running time $O(n^{(q+4)/(q+2)})$.

A splitting of an arbitrary vector 
$Z = \left(
\begin{array}{c}
Z_u\\
Z_v
\end{array}
\right)\in\RR^{m+3}$ into two vectors $Z_u\in\RR^m$ and $Z_v\in\RR^3$ implies the corresponding partitions 
\begin{align}
	g_{GSE}(y) &= 
\left(\begin{array}{c}
g_{GSE,u}(y)\\
g_{GSE,v}(y)
\end{array}
\right), \label{eq17}
\\
	G_{GSE,g}(y) &= 
\left(\begin{array}{c}
G_{GSE,g,u}(y)\\
G_{GSE,g,v}(y)
\end{array}
\right) \label{eq1711}
\end{align}    
of the mapping $g_{GSE}(y)$ and the matrix $G_{GSE,g}(y)$. 

Using the representation $Z = F(\mbtheta)$ \eqref{eq11}, the embedding mapping $y = h_{GSE}(Z)$, defined on the RM $\mbM$, can be written as a function
\begin{equation}
\label{eq18}
	y = R_{GSE}(\mbtheta) \equiv h_{GSE}(F(\mbtheta)),
\end{equation}    
defined on the LS $\mbTheta$.

Using the mapping $\mbtheta = V(w)$, the RM $\mbM$ and the embedding mapping $y = h_{GSE}(Z)$ can be written as 
\[
	\mbM = \{Z=f(w),\,w\in\mathbf{W}\}\subset\RR^{m+3}
\]
and
\begin{equation}
\label{eq19}
	y = r_{GSE}(X) = h_{GSE}(f(w))
\end{equation}    
respectively, where the functions
\[
	f(w) = \left(\begin{array}{c}
w\\
V(w)
\end{array}
\right)
\]
and $r_{GSE}(w)$ \eqref{eq19} are defined on the VFS $\mbW$.

The $3\times m$ and $3\times3$ Jacobian matrices $J_{GSE,r}(w)$ (the covariant differentiation is used here) and $J_{GSE,R}(\mbtheta)$ of the mappings $r_{GSE}(w)$ and $R_{GSE}(\mbtheta)$ can be estimated [20] by the matrices
\begin{align}
\label{eq20}
	G_{GSE,r}(w) &= G_{GSE,g,u}^- (r_{GSE}(w)) \times \pi_{GSE,\mbX}(w),	\\
    \label{eq21}
	G_{GSE,R}(\mbtheta) &= G_{GSE,g,v}^{-1} (R_{GSE}(\mbtheta)),
\end{align}    
respectively. Here $H^{-} = (H^{\mathrm{T}}\times H)^{-1}\times H^{\mathrm{T}}$ denotes a pseudo-inverse Moore-Penrose matrix \cite{bib28} of an arbitrary matrix $H$ and $\pi_{GSE,\mbX}(w)$ is a certain estimator \cite{bib20} of an $m\times m$ projection matrix onto the tangent space of the VFS $\mbW$ at the point $w \in \mbW$.

Using representations \eqref{eq18} and \eqref{eq19}, the proximity \eqref{eq14} implies  approximate equalities
\begin{align}
\label{eq22}
	V_{GSE}(w) &\equiv g_{GSE,v}(r_{GSE}(w)) \approx V(w),\\
    \label{eq23}
	\Phi_{GSE}(\mbtheta) &\equiv g_{GSE,u}(R_{GSE}(\mbtheta)) \approx \Phi(\mbtheta). 	
\end{align}    

Although the GSE-based functions $V_{GSE}(w)$ \eqref{eq22} and $\Phi_{GSE}(\mbtheta)$ \eqref{eq23} accurately approximate the sought-for functions $\psi(w)$ and $\Phi(\mbtheta)$, respectively, they cannot be considered as the solution to the Robot Localization problem because the mappings $g_{GSE,u}(y)$ and $g_{GSE,v}(y)$ \eqref{eq17}, \eqref{eq1711} depend on the argument 
\begin{equation}
\label{eq24}
	y = r_{GSE}(w) = R_{GSE}(\mbtheta),
\end{equation}    
whose values are known only at sample points:
\begin{equation}
\label{eq25}
	y_i = h_{GSE}(Z_i) = r_{GSE}(w_i) = R_{GSE}(\mbtheta_i), i = 1, 2, \ldots , n.
\end{equation}    

Based on known values \eqref{eq25} of the functions $r_{GSE}(w)$ \eqref{eq19} and $R_{GSE}(\mbtheta)$ \eqref{eq18} at sample points, as well as on the known values of their Jacobian matrices \eqref{eq20}, \eqref{eq21} at these points, the estimators $r^*(w)$ and $R^*(\mbtheta)$ of these functions at arbitrary points $w \in \mbW$ and $\mbtheta\in\mbTheta$, respectively, are constructed using the Jacobian Regression method, proposed in \cite{bib29}. 

GSE solution applied to the RM $\mbM$ includes construction of the kernels $K_{\mathbf{W}}(w, w')$ and $K_{\mbTheta}(\mbtheta, \mbtheta')$ on the VFS $\mbW$ and the LS $\mbTheta$, respectively. These kernels reflect not only geometrical closeness between points $Z = F(\mbtheta) = f(X)$ and $Z' = F(\mbtheta') = f(X')$ but also closeness between the tangent spaces $L(Z)$ and $L(Z')$ to the RM $\mbM$. 

Using these kernels, the Jacobian Regression method gives the estimators $r^*(w)$ and $R^*(\mbtheta)$ in an explicit form defined by formulas

\[
	\frac{1}{K_{\mathbf{W}}(w)} \sum_{i=1}^n K_{\mathbf{W}}(w,w_i)\times\left\{y_i+G_{GSE,r}(w_i)\times(w-w_i)\right\}
\]
and
\[
	\frac{1}{K_{\mbTheta}(\mbtheta)} \sum_{i=1}^n K_{\mbTheta}(\mbtheta, \mbtheta_i)\times\left\{y_i+G_{GSE,R}(\mbtheta_i)\times(\mbtheta-\mbtheta_i)\right\}
\]
respectively, where
$K_{\mathbf{W}}(w) = \sum_{i=1}^n K_{\mathbf{W}}(w,w_i)$ and 
$K_\mbTheta(\mbtheta) = \sum_{i=1}^n K_{\mbTheta}(\mbtheta,\mbtheta_i)$.

\subsection{Robot Localization: Final Formulas}

Let us denote $g_{\mbTheta}(y) = g_{GSE,v}(y)$ and $g_{\mathbf{W}}(y) = g_{GSE,u}(y)$. Substitution of estimators $r^*(w)$ and $R^*(\mbtheta)$ in formulas \eqref{eq22} and \eqref{eq23} instead of $r_{GSE}(w)$ and $R_{GSE}(\mbtheta)$, provides the final estimator 
\[
	\psi^*(X) = g_{\mbTheta}(r^*(\omega(X)))
\]
for the Localization function $\psi(X) = V(\omega(X))$, and the final estimator
\[
	\Phi^*(\mbtheta) \equiv g_{\mathbf{W}}(R^*(\mbtheta))
\]
for the Feature modeling function $w = \Phi(\mbtheta)$.

Note that most of known appearance-based learning methods solve only the Localization function estimation problem.

\section{Numerical experiments}
\label{sec5}

We used vehicle moving in a city ``Multi-FoV'' synthetic dataset \cite{Zhang2016} and considered it as a testing problem for robot localization methods. The data set consists of $n = 2500$ color images of size $480 \times 640$ and information about the vehicle position for each of them. Measurements from three different optical vision systems (perspective, fisheye and catadioptric) are presented in the dataset.

\begin{figure}[h]

  \begin{minipage}{.33\linewidth}
  \centering
  \subfloat[]{\label{main:a}\includegraphics[width=1\textwidth]{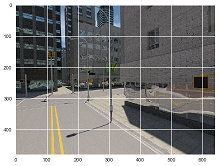}}
  \end{minipage}%
  \begin{minipage}{.33\linewidth}
  \centering
  \subfloat[]{\label{main:b}\includegraphics[width=1\textwidth]{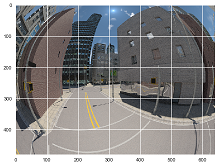}}
  \end{minipage}
  \begin{minipage}{.33\linewidth}
  \centering
  \subfloat[]{\label{main:c}\includegraphics[width=1\textwidth]{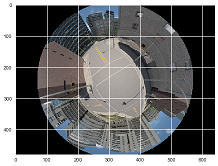}}
  \end{minipage}

  \caption{Examples of images from dataset: perspective (a), fisheye (b) and catadioptric (c)}
  \label{fig:1}
\end{figure}

Firstly, we estimated the intrinsic dimensionality of the data using three different popular approaches \cite{Camastra2016}: global via IsoMap \cite{Tenenbaum2000}, local via correlation dimension \cite{Granata2016} and pointwise via maximum likelihood method from \cite{Levina2004}.  The results of estimation are presented in Table \ref{table1}. The true dataset intrinsic dimension is equal to one by design as a dimensionality of a well-sampled continuous trajectory of the vehicle. The overestimated dimensionality by local and pointwise approaches could be caused by extremely large data dimensionality ($p\sim 10^6$).

\begin{table}[htbp]
\caption{Dataset dimensionality estimation}
\begin{center}
\begin{tabular}{|c|c|c|c|}
\hline
     vision system$\setminus$method        & global & local & pointwise \\ \hline
perspective  & 1      & 11    & 3        \\ \hline
fisheye      & 1      & 7     & 3        \\ \hline
catadioptric & 1      & 11    & 3        \\ \hline
\end{tabular}
\label{table1}
\end{center}
\end{table}

We also applied GSE-approach for localization vs kernel nonparametric regression (KNR). Random subsample with $70\%$ of data points was used as a training set, and the rest of data points was used as a testing set. Orientation was not considered. The optimal kernel bandwidth for the KNR was estimated globally and the same for all coordinates (color channels of pixels) using leave-one-out cross-validation. The relative root means square error (RRMSE) was used as an error measure. Results are presented in Table \ref{table2}. One can see that GSE provides better results. One of explanations is that GSE uses adaptive kernel bandwidth in the original space and takes into account first order effects.

\begin{table}[htbp]
\caption{RRMSE for localization problem}
\begin{center}
\begin{tabular}{|l|l|l|}
\hline
      vision system$\setminus$method       & GSE   & KNR   \\ \hline
perspective  & 0.045 & 0.063 \\ \hline
fisheye      & 0.037 & 0.059 \\ \hline
catadioptric & 0.039 & 0.058 \\ \hline
\end{tabular}
\label{table2}
\end{center}
\end{table}

\begin{figure}[h]

  \begin{minipage}{.33\linewidth}
  \centering
  \subfloat[]{\label{main:a}\includegraphics[width=1\textwidth]{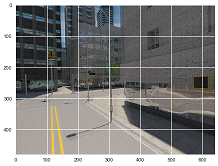}}
  \end{minipage}%
  \begin{minipage}{.33\linewidth}
  \centering
  \subfloat[]{\label{main:b}\includegraphics[width=1\textwidth]{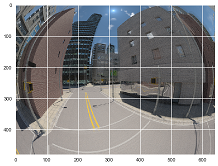}}
  \end{minipage}
  \begin{minipage}{.33\linewidth}
  \centering
  \subfloat[]{\label{main:c}\includegraphics[width=1\textwidth]{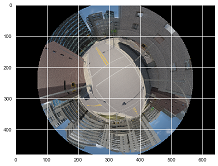}}
  \end{minipage}

  \caption{Examples of reduction-reconstruction mappings for out-of-sample image: perspective (a), fisheye (b) and catadioptric (c)}
  \label{fig:2}
\end{figure}

\begin{figure}[h]

  \begin{minipage}{.33\linewidth}
  \centering
  \subfloat[]{\label{main:a}\includegraphics[width=1\textwidth]{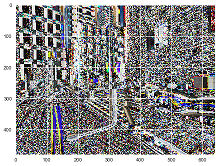}}
  \end{minipage}%
  \begin{minipage}{.33\linewidth}
  \centering
  \subfloat[]{\label{main:b}\includegraphics[width=1\textwidth]{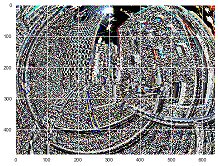}}
  \end{minipage}
  \begin{minipage}{.33\linewidth}
  \centering
  \subfloat[]{\label{main:c}\includegraphics[width=1\textwidth]{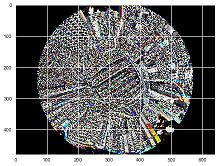}}
  \end{minipage}

  \caption{Examples of tangent to out-of-sample image: perspective (a), fisheye (b) and catadioptric (c)}
  \label{fig:3}
\end{figure}

The GSE-algorithm not only reduces dimensionality but also constructs reconstruction mapping (Figure \ref{fig:2}) and estimates differential structure such as tangent spaces (Figure \ref{fig:3}) which could be useful for velocity estimation. One can see that reduced-reconstructed images are almost the same as original ones but a bit blurred (the norm of difference $\sim 0.01$ of the original image Frobenius norm) and the tangent space represents borders of moving objects.

\section{Conclusions}
\label{sec6}

 We consider an appearance-based robot self-localization problem. Using recent manifold learning and deep learning techniques, we propose a new geometrically motivated solution based on training data consisting of a finite set of images captured in known locations of the robot. Numerical experiments demonstrated efficiency of the proposed approach. Further full-scale experiments under different external conditions are underway.

\vspace{1em}
\noindent\textbf{Acknowledgments}. 

E. Burnaev was supported by the RFBR grants 16-01-00576 A and 16-29-09649 ofi\_m. A. Bernstein and Y. Yanovich were supported by the Russian Science Foundation grant (project 14-50-00150).

\end{document}